\documentclass[default,iicol]{sn-jnl}

\usepackage{diagbox}
\usepackage{tabularx}

\jyear{2021}%

\theoremstyle{thmstyleone}%
\theoremstyle{thmstyletwo}%

\theoremstyle{thmstylethree}%

\begin{document}

\title[Pore-scale Patch Dataset and A Comparative Evaluation]{Pore-scale Image Patch Dataset and A Comparative Evaluation of Pore-scale Facial Features}


\author[1]{\fnm{Dong}   \sur{Li}\email{dongli@gdut.edu.cn}}          
\author[1]{\fnm{Hualiang} \sur{Lin}\email{2112304356@gdut.edu.cn}}    
\author[1]{\fnm{JiaYu} \sur{Li}\email{2112404090@gdut.edu.cn}}
\affil[1]{\orgdiv{School of Automation}, \orgname{Guangdong University of Technology}, \orgaddress{\city{Guangzhou} \postcode{510006}, \country{China}}}

\abstract{The weak-texture nature of facial skin regions presents significant challenges for local descriptor matching in applications such as facial motion analysis and 3D face reconstruction.
Although deep learning-based descriptors have demonstrated superior performance to traditional hand-crafted descriptors in many applications, the scarcity of pore-scale image patch datasets has hindered their further development in the facial domain. In this paper, we propose the PorePatch dataset, a high-quality pore-scale image patch dataset, and establish a rational evaluation benchmark. We introduce a Data-Model Co-Evolution (DMCE) framework to generate a progressively refined, high-quality dataset from high-resolution facial images. We then train existing SOTA models on our dataset and conduct extensive experiments. Our results show that the SOTA model achieves a FPR95 value of 1.91$\%$ on the matching task, outperforming PSIFT (22.41$\%$) by a margin of 20.5$\%$. However, its advantage is diminished in the 3D reconstruction task, where its overall performance is not significantly better than that of traditional descriptors. This indicates that deep learning descriptors still have limitations in addressing the challenges of facial weak-texture regions, and much work remains to be done in this field.}




\maketitle
\section{Introduction}\label{sec1}
Local Feature Descriptors are a fundamental technology in computer vision, dedicated to encoding the neighborhood information of keypoints into compact and discriminative feature vectors. Due to their robustness to geometric and photometric transformations, these descriptors play an indispensable role in numerous downstream tasks, including image matching \cite{ma2021image}, image retrieval \cite{dubey2021decade}, Structure from Motion (SfM) \cite{shalaby2017algorithms}, and Simultaneous Localization and Mapping (SLAM) \cite{kazerouni2022survey}.
Over the past two decades, the development of local descriptors has undergone a paradigm shift from handcrafted design to deep learning. Traditional handcrafted descriptors, represented by methods such as SIFT \cite{lowe2004distinctive}, SURF \cite{bay2008speeded}, and ORB \cite{rublee2011orb}, achieved great success by constructing features from meticulously designed statistical information like gradients and orientations. Subsequently, with the rise of deep learning, descriptors based on Convolutional Neural Networks (CNNs), such as HardNet \cite{mishchuk2017working}, HyNet \cite{tian2020hynet}, and SDGMNet \cite{deng2024sdgmnet}, have demonstrated superior performance by leveraging end-to-end metric learning on large-scale image patch datasets \cite{brown2010discriminative} to automatically learn feature representations that are more robust and powerful than their handcrafted counterparts \cite{joshi2020recent,xu2024local} across standard benchmarks \cite{brown2010discriminative,balntas2017hpatches,schonberger2017comparative}.

However, when the application scenario shifts from conventional rigid, texture-rich objects (e.g., buildings, landscapes) to human faces, existing descriptor methods face severe challenges. Facial skin regions typically exhibit weak-texture characteristics, lacking distinct corners or edges, which makes it difficult for traditional descriptors to capture stable and discriminative features \cite{li2015design}. Although some studies have attempted to optimize traditional descriptors for facial characteristics \cite{lin2010accurate,li2015design}, the performance improvements have been relatively limited.
More critically, while deep learning descriptors have achieved significant success in other domains, their research in the facial domain has been hindered by a lack of large-scale, high-quality image patch datasets specifically designed for pore-scale facial features (e.g., at the pore-scale). Most existing deep learning descriptors are trained on general-purpose datasets, such as that from UBC PhotoTour \cite{brown2010discriminative}, which is created from building images from various locations. The features they learn are not fully applicable to the unique weak-texture and non-rigid deformation properties of faces. 
The scarcity of facial image patch datasets has become a core bottleneck constraining the development of more advanced descriptor algorithms in face-related domains.

To fill this void and advance the research and development of local descriptors in face-related domains, this paper proposes a comprehensive solution. We propose a novel method to construct PorePatch, a large-scale dataset of pore-scale face patches
More importantly, we move beyond a one-time data processing by proposing a Data-Model Co-Evolution (DMCE) framework, which continuously improves the dataset's quality and challenge through a performance-driven iterative loop.
Finally, we conduct a comprehensive benchmark to evaluate state-of-the-art descriptors, assessing their performance in patch verification and the downstream task of 3D reconstruction. In summary, the main contributions of our work are as follows:
\begin{enumerate}
\renewcommand{\theenumi}{\arabic{enumi})}
\item We propose a novel method for constructing a large-scale, pore-scale face patch dataset, named the PorePatch dataset, to address the critical lack of specialized training data for learning local descriptors on weak-textured facial regions.

\item We introduce a Data-Model Co-Evolution (DMCE) framework, an iterative refinement strategy that mutually enhances data quality and model performance. This approach proved highly effective, culminating in a final dataset ($PorePatchV_4$) with 95.1$\%$ more patches than the initial version ($PorePatchV_1$). This substantial improvement in data quality directly translated to superior model performance, with the top-performing descriptor (AFSRNet) achieving a 14.7$\%$ lower FPR95 error rate on the final dataset compared to the first.

\item We provide a comprehensive benchmark of state-of-the-art descriptors on our PorePatch dataset, with performance assessed based on patch verification metrics and the quality of 3D reconstruction.
\end{enumerate}

\begin{table*}
  \caption{Comparison of dataset properties. ``Human Face''  indicates if the dataset consists of facial images. ``Non-Planar Scene'' denotes whether scenes with 3D parallax are included. ``Ground Truth Generation'' specifies the method used to obtain patch correspondences. ``Approximate Patch Number'' is the total count of patches in the dataset.}
  \label{tab:dataset property}
  \centering
  \small
  \setlength{\tabcolsep}{4pt}
  \footnotesize
  \begin{tabular}{lcccc}
    \hline
    \textbf{Dataset} & \textbf{Human Face} & \textbf{Non-Planar Scene} & \textbf{Ground Truth Generation} & \textbf{Approximate Patch Number} \\
    \hline
    UBC PhotoTour \cite{brown2010discriminative} & & \checkmark & SfM + MVS & 1,200,000 \\
    Synthetic \cite{fischer5769descriptor}     & & & Homography & 2,400,000 \\
    PhotoSynth \cite{mitra2018large}    & & \checkmark & SfM  & 7,500,000 \\
    PorePatch (proposed) & \checkmark & \checkmark & SfM + MVS  + DMCE & 10,000,000 \\
    \hline
  \end{tabular}
\end{table*}
\section{Related Work}\label{sec2}
\subsection{Local Feature Descriptors}
Early research has primarily focused on handcrafted descriptors, which required specialized engineering knowledge \cite{ma2021image}. SIFT \cite{lowe2004distinctive} is widely regarded as the most prevalent descriptor, which computes a smoothed histogram from the gradient field of an image patch. Subsequent methods, such as PCA-SIFT \cite{ke2004pca}, were developed to improve upon SIFT's robustness by applying techniques like Principal Component Analysis. Furthermore, some research has focused on adapting these ideas for specific objects such as human faces \cite{lin2010accurate,li2015design}. To address the matching problem in weak-textured regions of facial skin, Pore-SIFT(PSIFT) \cite{li2015design} was proposed. The core idea of this method is that while individual pore-scale features are not distinctive, the spatial layout of neighboring feature points is stable; therefore, it first detects keypoints at the pore-scale and then encodes this relative positional information into the descriptor by expanding SIFT's neighborhood support region. Although these handcrafted descriptors achieved significant success, their design is fundamentally based on low-level visual cues, such as gradient orientation histograms or pairwise intensity comparisons.

Recently, propelled by the tremendous success of deep learning in computer vision, patch-based deep learning descriptors have undergone rapid development.
L2Net \cite{tian2017l2} introduces a widely adopted network to replace the Siamese architecture, further improving performance. HardNet \cite{mishchuk2017working}, leveraging the L2Net architecture, employs a simple yet effective strategy known as hard negative mining, highlighting the importance of proper sampling. RALNet \cite{xu2019robust} measures similarity using the angular distance between feature vectors instead of the L2 distance. HyNet \cite{tian2020hynet} introduces a hybrid similarity metric loss, combining L2 distance with cosine similarity, which improved performance across different datasets. CDF-STC \cite{yin2022stcdesc} considers the correlation between descriptors from different types of samples and proposes a descriptor network based on a triplet loss with a similar triangle constraint. SDGMNet \cite{deng2024sdgmnet} uses a statistics-based autofocus modulation method to mitigate the impact of extremely hard samples, leading to more stable training convergence. AFSRNet \cite{li2024afsrnet} utilizes a multi-scale network to integrate richer information and incorporates a novel regularization loss based on the symmetry of the similarity matrix to improve feature robustness.

\begin{table*}[htbp]
\caption{Statistics of the iterative dataset generation process. This table provides a comprehensive summary of each dataset version, including the number of successful and failed subjects, the total count of dense points, and the distribution of patches for both training and testing sets. The 'Generating Descriptor' column indicates the specific descriptor used to generate each version of the dataset.}
\label{tab:my_task_data}
\centering
\small
\setlength{\tabcolsep}{4pt}
\begin{tabular}{l c c c c c c c}
\toprule
\textbf{Dataset} &\textbf{Generating} &\textbf{Successful} & \textbf{Failed} & \textbf{Dense} & \textbf{Total} & \textbf{Train} & \textbf{Test} \\
\textbf{Version} &\textbf{Descriptor} &\textbf{Subjects} & \textbf{Subjects} & \textbf{Points} & \textbf{Patches} & \textbf{Patches} & \textbf{Patches} \\
\midrule
$PorePatchV_1$ &PSIFT  &117 & 140 & 568,825 & 5,137,273 & 3,479,727 & 1,657,546 \\
$PorePatchV_2$ &$Pore-HyNet-V_1$ &213 & 44 & 888,981 & 7,882,260 & 6,225,657 & 1,656,603 \\
$PorePatchV_3$ &$Pore-AFSRNet-V_2$ &257 & 0 & 1,103,749 & 9,842,654 & 8,184,149 & 1,658,505 \\
$PorePatchV_4$ &$Pore-AFSRNet-V_3$ &257 & 0 & 1,105,594 & 10,024,242 & 8,359,278 & 1,664,964 \\
\bottomrule
\end{tabular}
\end{table*}
\subsection{Datasets for Local Descriptor Learning}
The success of deep learning-based descriptors is largely attributed to the emergence of large-scale, high-quality image patch datasets. Table \ref{tab:dataset property} provides a comparative overview of datasets in this domain. The UBC PhotoTour dataset \cite{brown2010discriminative} represents a foundational work in this direction. This dataset is created by applying Structure from Motion (SfM) to tourist photographs to reconstruct real-world 3D scenes. Subsequently, Multi-View Stereo (MVS) techniques are utilized to obtain denser surface models, yielding a collection of image patches with ground-truth geometric correspondences. This technological advance leads to a larger-scale benchmark, comprising three scenes with over 400,000 image patches each, that exhibits rich variations in viewpoint, illumination, and geometry.
The Synthetic dataset~\cite{fischer5769descriptor} is generated by randomly sampling 16,000 patches from images and applying 150 random transformations to each. Although this method can produce a virtually unlimited number of matching pairs, its underlying transformation model is typically restricted to planar homographies, thus failing to fully replicate the complex geometric effects caused by real-world 3D parallax.
To address the limitations of the prevalent UBC PhotoTour dataset, such as its limited number of scenes and lack of diversity, PhotoSynth~\cite{mitra2018large} provides a larger and more extensive image patch dataset, constructed using Structure from Motion. It features richer variations in real-world viewpoint, illumination, and scale. The dataset comprises a total of 30 scenes, with an average of 250,000 patches extracted from each. The introduction of this dataset advances the matching performance of descriptors on standard benchmarks, particularly in challenging wide-baseline scenarios.
Nevertheless, many contemporary descriptors are still trained and studied on the UBC PhotoTour dataset. This is largely due to its high-quality ground-truth correspondences and its established role as a canonical benchmark, which ensures fair comparisons with prior methods. Therefore, the creation of our dataset is modeled after the UBC PhotoTour pipeline.

Developing descriptors through specialized datasets is a proven and effective paradigm, particularly in challenging domains such as cross-modal remote sensing \cite{yu2023feature, xu2023sar}. These efforts highlight a critical trend: to boost descriptor performance under unique image characteristics and application demands, constructing domain-specific, high-quality training datasets is a crucial step. However, despite the face being a central research object in computer vision, a dedicated, high-quality patch dataset for learning local face descriptors remains, to our knowledge, a significant gap.

To fill this void, we introduce the PorePatch dataset, a large-scale, multi-view patch dataset specifically designed for pore-scale facial analysis, constructed from the high-definition Nersemble dataset \cite{kirschstein2023nersemble}. As summarized in Table~\ref{tab:dataset property}, PorePatch distinguishes itself from existing datasets in several key aspects. Firstly, it directly addresses the scarcity of specialized data for faces, a domain not covered by influential general-purpose datasets like UBC PhotoTour and PhotoSynth. Secondly, with approximately 10 million patches, it significantly surpasses previous datasets in scale. Most importantly, while leveraging an SfM and MVS pipeline to capture realistic non-planar geometry, PorePatch pioneers a novel Data-Model Co-Evolution (DMCE) framework for ground truth generation, which introduces a performance-driven iterative loop to generate progressively higher-fidelity 3D reconstructions. This process in turn yields a dataset of superior quality and complexity, facilitating the training of more robust descriptors.

\section{PorePatch dataset}\label{sec3}
This section details various aspects of the PorePatch dataset. This paper introduces a high-quality face patch dataset generated via our novel Data-Model Co-Evolution (DMCE) framework. It is named the PorePatch dataset because it is constructed based on keypoints detected at the pore-scale using PSIFT \cite{li2015design}. Section \ref{DoftPd} describes the details of the PorePatch dataset, Section \ref{CoftPd} explains its creation pipeline, and Section \ref{IOoPd} outlines the DMCE methodology.

\subsection{Description of the PorePatch dataset}\label{DoftPd}
For our PorePatch dataset, which necessitates high-resolution images of human faces captured from multiple distinct viewpoints, we opt for the NeRsemble dataset \cite{kirschstein2023nersemble}. NeRsemble stands out as a large-scale database for high-resolution, multi-view sequences of human faces. It provides synchronous images from a 16-camera array at a 7.1 megapixel resolution, which serves as the input for our dataset generation framework. 
\begin{figure}[htbp]
    \centering
    \includegraphics[width=\columnwidth]{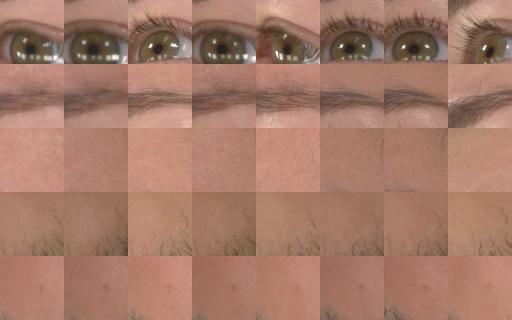}
    \caption{Sample patches from the PorePatch dataset. These patches are extracted around pore-scale keypoints and exhibit challenging weak-texture patterns. Patches are considered a match if they correspond to the same 3D point.}
    \label{fig:patchdatasetshow}
\end{figure}

The construction of the PorePatch dataset was driven by our DMCE framework, which facilitated its evolution through four progressively refined versions, from $PorePatchV_1$ to $PorePatchV_4$. For brevity, the term ``PorePatch dataset'' in this paper refers to the final and highest-quality version, $PorePatchV_4$, unless otherwise specified. The final dataset contains 257 unique subject identities, with an additional 9 distinct identities held out exclusively for subsequent 3D reconstruction experiments. To ensure the model's generalization capability, the dataset was curated to include subjects of diverse skin tones and ethnic backgrounds. We process 16 multi-view images for each subject, yielding an average of 37,000 patches per subject and a total of approximately 10 million patches. Examples of the pore-scale image patches generated by our framework are shown in Figure \ref{fig:patchdatasetshow}. The dataset is partitioned into training and test sets as detailed in Table \ref{tab:my_task_data}.
\begin{algorithm*}[htbp]
\caption{The DMCE Framework for Dataset Generation}
\label{alg:iterative_framework}
\begin{algorithmic}[1]
    \Require
        Set of multi-view images for all subjects \( I \);
        Initial handcrafted descriptor algorithm \( D_{PSIFT} \);
        Set of candidate deep learning architectures \( A = \{A_1, A_2, \dots\} \);
        Total number of iterations \( N_{iter} \).
    \Ensure
        A sequence of iteratively refined datasets \( \{\text{PorePatchV}_1, \dots, \text{PorePatchV}_{N_{iter}}\} \).

    \Statex
    \Function{ConstructPatchDataset}{$I, \text{Model}$}
        \State \( (\mathbf{P}_{\text{dense}}, \{C_i\}) \gets \text{Reconstruct3D}(I, \text{Model}) \) \Comment{Generates dense point cloud \(\mathbf{P}_{\text{dense}}\) and camera parameters \(\{C_i\}\) via the SfM+MVS pipeline}
        \State \( \mathcal{K} \gets \text{DetectKeypoints}(I_{\text{frontal}}) \) \Comment{Detect keypoint set \(\mathcal{K}\) on the frontal-view image}
        \State \( \mathbf{P}_{\text{kpt-anc}} \gets \{ p \in \mathbf{P}_{\text{dense}} \mid \min_{kpt \in \mathcal{K}} \| \mathcal{P}(p, C_{\text{frontal}}) - kpt \| \le 5 \} \)
        \State \( \text{PatchDict} \gets \text{Initialize empty dictionary} \)
        \For{each point \( p_k \in \mathbf{P}_{\text{kpt-anc}} \)}
            \State \( \text{PatchDict}[p_k] \gets \text{ProjectAndExtract}(p_k, I, \{C_i\}) \)
        \EndFor
        \State \( (\text{Patch}_{\text{pos}}, \text{Patch}_{\text{neg}}) \gets \text{GeneratePairs}(\text{PatchDict}) \)
        \State \Return \( \{\text{Patch}_{\text{pos}}, \text{Patch}_{\text{neg}}\} \)
    \EndFunction
    \Statex

    \Statex \textit{// Main iterative process}
    \Statex \textit{// --- Bootstrapping Phase ---}
    \State \( \text{PorePatchV}_1 \gets \text{ConstructPatchDataset}(I, D_{PSIFT}) \) \Comment{Generate V1 using PSIFT}
    
    \Statex \textit{// --- Iterative Refinement Phase ---}
    \For{\( n \gets 2 \) to \( N_{iter} \)}
        \State \( \text{TrainedModels} \gets \emptyset \)
        \For{each architecture \( A_j \in A \)}
            \State \( \text{Pore-}A_j\text{-V}_{n-1} \gets \text{Train}(A_j, \text{PorePatchV}_{n-1}) \) \Comment{Train candidates}
            \State Add \( \text{Pore-}A_j\text{-V}_{n-1} \) to \( \text{TrainedModels} \)
        \EndFor
        \State \( \text{BestModel} \gets \text{EvaluateAndSelectBest}(\text{TrainedModels}) \) \Comment{Select the best model}
        \State \( \text{PorePatchV}_n \gets \text{ConstructPatchDataset}(I, \text{BestModel}) \) \Comment{Generate new dataset}
    \EndFor
    
    \State \Return \( \{\text{PorePatchV}_1, \dots, \text{PorePatchV}_{N_{iter}}\} \)
\end{algorithmic}
\end{algorithm*}
\subsection{The Dataset Generation Pipeline}\label{CoftPd}
The entire framework for generating the PorePatch dataset is formally outlined in Algorithm \ref{alg:iterative_framework}. This process comprises two main stages: an initial patch dataset construction phase, detailed in this section, and an DMCE phase, explained in Section \ref{IOoPd}. The core of our framework, detailed in the \textsc{ConstructPatchDataset} function in Algorithm \ref{alg:iterative_framework} (lines 1-11), begins with the generation of ground-truth correspondences using Structure from Motion (SfM). The primary output of the SfM process is a 3D point cloud of the scene. Each point in this cloud corresponds to a list of 2D feature points that have been matched across images from different viewpoints. Using these feature correspondences established across multiple images, SfM employs triangulation to calculate the precise 3D positions of these points. It also estimates the camera parameters for each image, including intrinsics and extrinsics. In our reconstruction pipeline, we use the PSIFT method to detect keypoints. These keypoints are then described by a local descriptor algorithm to generate feature vectors. The correspondences established by matching these features across multiple views serve as the initial input for 3D reconstruction~\cite{schoenberger2016sfm,schoenberger2016mvs}.

To generate the dense point cloud required for the subsequent patch association, our pipeline builds upon the initial SfM output by employing Multi-View Stereo (MVS). The MVS stage leverages the previously estimated camera parameters and the input images to perform a dense reconstruction. This is achieved by matching a large number of pixels across multiple views, which results in a significantly more complete and detailed dense point cloud than the initial one generated from only discrete feature points.
To associate the dense point cloud with 2D feature points, we reproject the 3D points onto the frontal view and match them with the keypoints detected on that view via the PSIFT method. A 3D point is labeled as an ``keypoint-anchored 3D point'' if and only if the distance between its projection in the frontal view and the nearest 2D keypoint is within a preset threshold of 5 pixels.

Subsequently, we reproject these keypoint-anchored 3D points onto other image views to obtain their corresponding image patches.
Based on the projected positions of these points in their respective views, we extract local 64x64 pixel image patches centered at these locations.
This patch size was chosen primarily to facilitate subsequent data augmentation (e.g., rotation, homography transformations) and to ensure compatibility with various common local descriptor algorithms.
When constructing the patch pairs for descriptor learning, we define patches originating from the same 3D point in different views as a positive pair, while patches originating from different 3D points are defined as a negative pair.

\subsection{Data-Model Co-Evolution for Dataset Generation}\label{IOoPd}
The generation of the PorePatch dataset is guided by our DMCE framework, as depicted in the main performance-driven iterative loop of Algorithm \ref{alg:iterative_framework} (lines 13-21). Within this framework, data and model quality mutually reinforce each other: a higher-quality dataset is used to train more powerful descriptor models, which in turn are leveraged to generate an even better dataset for the next iteration. Specifically, the local descriptor driving the SfM and MVS pipeline is replaced by the top-performing model from the preceding stage. This enhancement leads to more accurate and complete 3D reconstructions, thereby increasing both the scale and geometric precision of the resulting patch dataset. This progressively refined data is crucial for enabling descriptor models to learn more robust and discriminative representations.

The iterative process begins with the initial version of the dataset, $PorePatchV_1$. This version was constructed using the handcrafted PSIFT descriptor, providing the necessary seed data for the subsequent learning-based methods. For each subsequent version, $PorePatchV_N$ (where $N>1$), we follow a ``train-evaluate-regenerate'' paradigm. Specifically, we first train a set of candidate deep learning descriptors on the preceding dataset, $PorePatchV_{N-1}$. These descriptors are then comprehensively evaluated according to the protocol detailed in section \ref{sec4}. The descriptor with the best overall performance on the 3D face reconstruction task is selected. This selected descriptor is subsequently used as the feature extractor to generate the higher-quality $PorePatchV_N$ dataset, while the remaining steps in the creation pipeline remain unchanged. This performance-driven iterative process is summarized in Algorithm \ref{alg:iterative_framework}

\section{Experiments}\label{sec4}

This section presents a comprehensive evaluation of the proposed PorePatch dataset and the performance of various local descriptors on pore-scale facial features. We benchmark the handcrafted PSIFT \cite{li2015design} against a suite of state-of-the-art deep learning descriptors: HardNet \cite{mishchuk2017working}, SOSNet \cite{tian2019sosnet}, RalNet \cite{xu2019robust}, HyNet \cite{tian2020hynet}, CDF-STC \cite{yin2022stcdesc}, SDGMNet \cite{deng2024sdgmnet}, and AFSRNet \cite{li2024afsrnet}. Their properties are detailed in Table \ref{tab:descriptor_properties}. Deep learning models trained on specific dataset versions are denoted as $Pore-Descr-V_i$. Due to computational constraints, the full iterative training (V1 through V4) is conducted only for the top-performing descriptors (HyNet, SDGMNet, and AFSRNet), while others are evaluated solely on $V_1$.

\begin{figure}[htbp]
    \centering
    \includegraphics[width=\columnwidth]{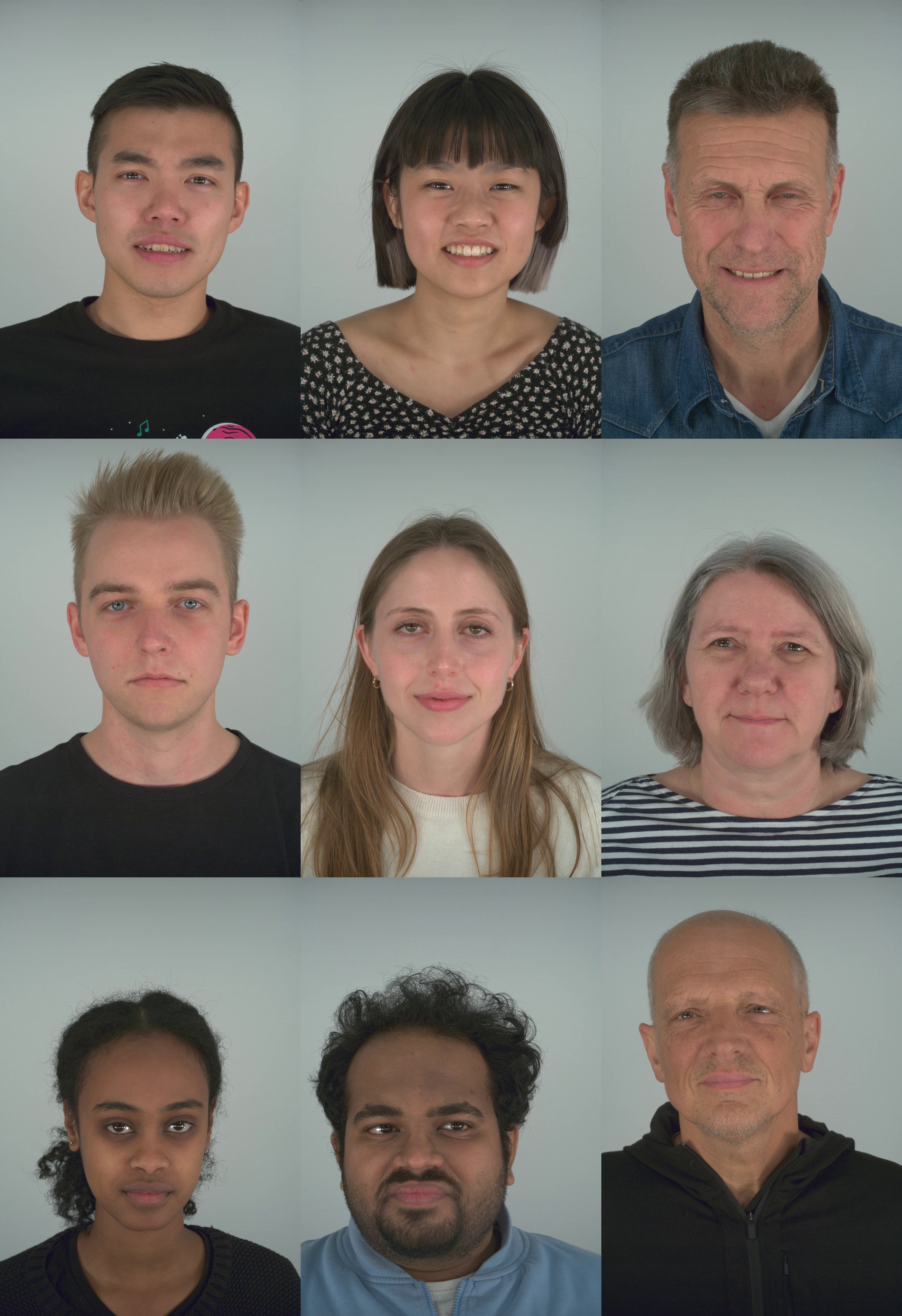}
    \caption{The 9 subjects used for the 3D face reconstruction task. In order from top to bottom, left to right, their respective IDs are: '031', '038', '124', '056', '097', '196', '175', '190', and '303'.}
    \label{fig:ethhuman}
\end{figure}

\begin{table*}[h]
\caption{Specifications of the local descriptors evaluated in this paper, including their vector dimensions, input patch sizes, and publication years.}\label{tab:descriptor_properties}
\centering
\small
\setlength{\tabcolsep}{4pt}
\begin{tabular}{@{}lcccccccc@{}}
\toprule
\textbf{Descr.} & \textbf{PSIFT} & \textbf{HardNet} & \textbf{SOSNet} & \textbf{RalNet} & \textbf{HyNet} & \textbf{CDF-STC} & \textbf{SDGMNet} & \textbf{AFSRNet} \\
\midrule
Dimensions    & 512    & 128    & 128    & 128    & 128    & 128    & 128    & 384 \\
Patch Size    & 64     & 32     & 32     & 32     & 32     & 32     & 32     & 64 \\
Publish Year  & 2015   & 2017   & 2019   & 2019   & 2020   & 2022   & 2024   & 2024 \\
\botrule
\end{tabular}
\end{table*}

\subsection{Data-Model Co-Evolution of the PorePatch Dataset}\label{sec:evolution}

\begin{table*}[htbp]
\caption{Average 3D reconstruction performance on all subjects across the iterative stages of the DMCE framework. The values represent the average metrics over the 9 held-out subjects used for the reconstruction task. The table shows the performance of descriptors trained on each version of the PorePatch dataset. The best performing value for each metric *within each stage* is highlighted in \textbf{bold}. The descriptor selected to generate the next dataset version is marked with (*).}
\label{tab:recon_evolution_avg}
\centering
\resizebox{\textwidth}{!}{%
\begin{tabular}{@{}l rrr rrr@{}}
\toprule
\textbf{Descriptor Method} & \textbf{Reg. Imgs} & \textbf{Sparse Pts} & \textbf{Track Len.} & \textbf{Reproj. Err.↓} & \textbf{Dense Pts} & \textbf{Inlier Matches} \\
\midrule
\multicolumn{7}{l}{\textit{Stage 1: Trained on PorePatchV$_1$}} \\
\midrule
PSIFT            & 16 & \textbf{7901.7} & 3.559 & 1.186 & 77,918.8 & \textbf{90,293.9} \\
$Pore-HardNet-V_1$  & 16 & 2,047.4 & 3.714 & 0.807 & 71,663.1 & 43,286.3 \\
$Pore-SOSNet-V_1$   & 16 & 1,926.8 & 3.712 & 0.824 & 72,748.1 & 36,492.6 \\
$Pore-RalNet-V_1$   & 16 & 2,050.7 & 3.757 & \textbf{0.792} & 72,209.7 & 39,779.3 \\
$Pore-HyNet-V_1$*   & 16 & 2,655.2 & 3.858 & 0.858 & \textbf{81,915.3} & 47,895.3 \\
$Pore-CDF-STC-V_1$  & 16 & 1,928.6 & 3.688 & 0.830 & 70,216.2 & 41,019.2 \\
$Pore-SDGMNet-V_1$ & 16 & 2,622.3 & \textbf{3.871} & 0.875 & 81,766.8 & 49,253.7 \\
$Pore-AFSRNet-V_1$ & 16 & 2,681.3 & 3.698 & 0.898 & 80,255.7 & 54,958.3 \\
\midrule
\multicolumn{7}{l}{\textit{Stage 2: Trained on PorePatchV$_2$}} \\
\midrule
$Pore-AFSRNet-V_2$* & 16 & \textbf{2938.7} & 3.766 & 0.926 & 83,469.8 & \textbf{51,211.0} \\
$Pore-HyNet-V_2$   & 16 & 2,683.3 & \textbf{3.850} & \textbf{0.879} & 81,282.6 & 45,211.0 \\
$Pore-SDGMNet-V_2$ & 16 & 2,782.7 & 3.847 & 0.916 & \textbf{85,140.4} & 47,897.8 \\
\midrule
\multicolumn{7}{l}{\textit{Stage 3: Trained on PorePatchV$_3$}} \\
\midrule
$Pore-AFSRNet-V_3$* & 16 & \textbf{3,234.2} & 3.809 & 0.939 & \textbf{88,631.1} & \textbf{55,140.0} \\
$Pore-HyNet-V_3$   & 16 & 2,762.8 & \textbf{3.866} & \textbf{0.884} & 84,034.2 & 50,081.6 \\
$Pore-SDGMNet-V_3$ & 16 & 2,739.3 & 3.833 & 0.903 & 84,475.4 & 47,341.7 \\
\bottomrule
\end{tabular}%
}
\end{table*}
For the 3D face reconstruction task, we adopt the evaluation protocol from the ETH benchmark \cite{schonberger2017comparative}, adapting it specifically for weak-textured facial surfaces. Performance is assessed using the standard metrics defined in the ETH protocol: Registered Images, Sparse Points, Track Length, Reprojection Error, Inlier Matches, and Dense Points. Following the ETH methodology, we implement both the Structure-from-Motion (SfM) and Multi-View Stereo (MVS) pipelines utilizing the COLMAP library \cite{schoenberger2016sfm,schoenberger2016mvs}. To drive the Data-Model Co-Evolution (DMCE), we employ a hierarchical selection protocol to identify the best descriptor for the next iteration. The primary criterion is the number of Dense Points, reflecting reconstruction completeness. In cases of comparable density (within 5\%), we prioritize Sparse Points and geometric accuracy metrics. To evaluate generalization capabilities, we selected a dedicated subset of 9 diverse subjects from the NeRsemble dataset (see Fig. \ref{fig:ethhuman}). Crucially, these subjects were strictly held out and excluded from the dataset generation pipeline. This section details the core iterative process of our DMCE framework, where the dataset evolves from $PorePatchV_1$ to $PorePatchV_3$. The selection of the best descriptor at each stage is guided by its quantitative performance on the 3D face reconstruction task. Table \ref{tab:recon_evolution_avg} summarizes the average performance on all subjects of descriptors across the three key iterative stages. The initial evaluation of descriptors trained on the $PorePatchV_1$ dataset, based on an aggregated assessment across all nine subjects as presented in the first section of Table \ref{tab:recon_evolution_avg}, reveals performance differences between the traditional PSIFT method and deep learning-based approaches. Although PSIFT shows an advantage in the number of sparse points and inlier matches, this is typically achieved at the cost of geometric accuracy. Its reprojection error, generally greater than 1.1 pixels and higher than all the deep learning methods, combined with its relatively short track lengths, indicates insufficient accuracy and multi-view stability in the reconstruction. In contrast, deep learning methods, represented by AFSRNet, SDGMNet, and HyNet, achieve longer track lengths and a greater number of dense points while maintaining a low reprojection error. This demonstrates their good potential for generating high-quality, high-accuracy 3D reconstructions. From these better-performing models, we followed our core methodology of ``iterating with the best performer'' and selected $Pore-HyNet-V_1$, the top overall performer in the V1 stage, to construct the $PorePatchV_2$ dataset. This choice was made not only for its good sparse reconstruction quality but also because it generated a considerable number of dense points, thereby achieving a desirable balance between the completeness and accuracy of the reconstruction.

As shown in the second stage of Table \ref{tab:recon_evolution_avg}, after being trained on the $PorePatchV_2$ dataset, the candidate descriptors exhibited an overall upward trend in the core metrics for reconstruction completeness—the number of Sparse Points and Dense Points. This indicates that improving the dataset quality effectively enhances the geometric completeness of the reconstructed models, which is corroborated by the growth in the number of Sparse Points and Dense Points. In this round, AFSRNet demonstrated the most significant performance growth, leading across all completeness-related metrics. Consequently, we selected $Pore-AFSRNet-V_2$ to generate the $PorePatchV_3$ dataset.

Finally, after training on the $PorePatchV_3$ dataset, the candidate descriptors once again showed an overall performance improvement, further validating our iterative strategy. In this evaluation round (third stage of Table \ref{tab:recon_evolution_avg}), AFSRNet was particularly outstanding, further consolidating its lead over the other candidates. The model maintained its top position in the number of Sparse Points and Dense Points. Based on its exceptional performance, we selected $Pore-AFSRNet-V_3$ to construct the final $PorePatchV_4$ dataset.

\subsection{Patch Verification on the Final PorePatch Dataset V4}
For patch verification, we report the False Positive Rate at 95\% True Positive Rate (FPR95) \cite{balntas2017hpatches}. We utilize the $PorePatchV_4$ dataset, which is partitioned into training and testing sets at an approximate 8:2 ratio. All verification experiments are reported on the test set. The detailed results are presented in Table \ref{tab:fpr95_performance}. As shown, all deep learning-based descriptors significantly outperform the traditional handcrafted descriptors. The classic SIFT descriptor performed the worst on this task, with a high FPR95 value of 37.61\%. PSIFT, which is specifically designed for facial features, outperformed SIFT with an FPR95 of 22.41\%, but it still lags far behind the learning-based methods. Even the most basic deep learning descriptor, $Pore-HardNet-V_1$ (3.92\%), achieved an error rate an order of magnitude lower than that of PSIFT. This clearly demonstrates the advantage of deep learning methods in extracting discriminative features from pore-scale face patches. Among the descriptors trained on $V_1$, $Pore-AFSRNet-V_1$ performed the best with an FPR95 of 2.24\%, followed closely by $Pore-SDGMNet-V_1$ (2.66\%) and $Pore-HyNet-V_1$ (2.72\%). Therefore, we selected these three highly competitive descriptors for iterative training and comparison in subsequent versions.

\begin{table}[htbp]
        \centering
        \caption{FPR95 performance of different descriptors on the $PorePatchV_4$ test set.}
        \label{tab:fpr95_performance}
        \begin{tabular}{lc}
            \toprule
            \textbf{Descriptor Method} & \textbf{FPR95 Value↓} \\
            \midrule
            SIFT	& 37.61 \\
            PSIFT	& 22.41 \\
            \midrule
            $Pore-HardNet-V_1$	& 3.92 \\
            $Pore-RalNet-V_1$	& 3.73 \\
            $Pore-CDF-STC-V_1$	& 3.12 \\
            $Pore-SOSNet-V_1$	& 3.04 \\
            \midrule
            $Pore-HyNet-V_1$		& 2.72 \\
            $Pore-HyNet-V_2$		& 3.33 \\
            $Pore-HyNet-V_3$		& 3.02 \\
            $Pore-HyNet-V_4$		& \textbf{2.44} \\
            \midrule
            $Pore-SDGMNet-V_1$	& 2.66 \\
            $Pore-SDGMNet-V_2$	& 3.61 \\
            $Pore-SDGMNet-V_3$	& 3.12 \\
            $Pore-SDGMNet-V_4$	& \textbf{2.37} \\
            \midrule
            $Pore-AFSRNet-V_1$	& 2.24 \\
            $Pore-AFSRNet-V_2$	& 2.98 \\
            $Pore-AFSRNet-V_3$	& 2.78 \\
            $Pore-AFSRNet-V_4$	& \textbf{1.91} \\
            \bottomrule
        \end{tabular}
\end{table}

The results in Table \ref{tab:fpr95_performance} also clearly validate the effectiveness of our proposed DMCE framework by showing the performance trend of AFSRNet, HyNet, and SDGMNet when trained on different dataset versions. A notable phenomenon is a slight, temporary drop in performance for all three descriptors when moving from the $V_1$ to the $V_2$ version. For instance, the FPR95 of AFSRNet increased from 2.24\% to 2.98\%. However, as the iteration continued, performance began to steadily recover after the descriptors were trained on the $PorePatchV_3$, which contained more effective information. Ultimately, all three descriptors achieved their respective best performances after being trained on the $PorePatchV_4$, which is the most informative and highest-quality version. The FPR95 values for $Pore-AFSRNet-V_4$, $Pore-SDGMNet-V_4$, and $Pore-HyNet-V_4$ reached 1.91\%, 2.37\%, and 2.44\%, respectively, which are the lowest values for each descriptor across all versions.

Regarding the observed ``dip-then-rise'' performance trend, we attribute its fundamental cause to the differing optimization objectives at various stages of DMCE. In the initial iterative stage (from $V_1$ to $V_2$), the primary objective is to maximize the dataset's scale and diversity. By employing a higher-performance deep learning model, we incorporated a large number of previously intractable, complex samples, making the data distribution of $V_2$ more challenging than that of $V_1$. Consequently, the temporary decline in the FPR95 metric is a direct reflection of the model's generalization adjustment to adapt to the significantly increased complexity of the learning task. In the subsequent iterative stages (from $V_3$ to $V_4$), the optimization objective shifts to enhancing the internal quality of the dataset, namely the geometric precision of the 3D reconstructions. More robust models generate more precise ground-truth correspondences, leading to a steady improvement in the label quality of the subsequent training sets ($V_3$ and $V_4$). As models are trained on this higher-quality data, their feature discrimination capabilities are enhanced, manifesting as a continuous rise in the performance metric. This ``dip, rise, and eventual optimum'' trend powerfully demonstrates that DMCE produces a superior PorePatch dataset, which in turn enables descriptors to learn more robust and generalizable features.
\begin{table*}[htbp]
    \centering
    \caption{3D reconstruction performance of elite descriptors trained on the $PorePatchV_4$. Performance gains have saturated at this stage. For each Subject ID and the final 'All Subjects' average, the best performing value for each metric is highlighted in \textbf{bold}. (BEST) denotes the descriptor with the best overall performance for the corresponding subject ID.}
    \label{tab:recon_v4}
    \resizebox{\textwidth}{!}{%
    \begin{tabular}{@{}ll rrr rrr@{}}
        \toprule
        \textbf{Subject ID} & \textbf{Descriptor Method} & \textbf{Registered Images} & \textbf{Sparse Points} & \textbf{Mean Track Length} & \textbf{Reproj. Error (px)} & \textbf{Dense Points} & \textbf{Inlier Matches} \\
        \cmidrule(r){1-2} \cmidrule(lr){3-3} \cmidrule(lr){4-4} \cmidrule(lr){5-5} \cmidrule(lr){6-6} \cmidrule(lr){7-7} \cmidrule(l){8-8}
        \multirow{3}{*}{\textbf{124}} 
        & $Pore-AFSRNet-V_4$ & 16 & \textbf{7,093} & \textbf{4.73} & 0.653 & 120,879 & 87,904 \\
        & $Pore-HyNet-V_4$(BEST)   & 16 & 6,842 & 4.70 & 0.635 & \textbf{122,515} & \textbf{105,901} \\
        & $Pore-SDGMNet-V_4$ & 16 & 5,840 & 4.65 & \textbf{0.612} & 111,187 & 68,012 \\
        \midrule
        \multirow{3}{*}{\textbf{175}} 
        & $Pore-AFSRNet-V_4$(BEST) & 16 & \textbf{658} & 2.75 & 1.233 & \textbf{43,377} & 31,092 \\
        & $Pore-HyNet-V_4$   & 16 & 610 & \textbf{2.87} & \textbf{1.145} & 43,067 & \textbf{34,185} \\
        & $Pore-SDGMNet-V_4$ & 16 & 426 & 2.78 & 1.176 & 33,355 & 32,808 \\
        \midrule
        \multirow{3}{*}{\textbf{190}} 
        & $Pore-AFSRNet-V_4$ & 16 & 1,638 & 3.29 & 0.888 & 89,214 & 55,389 \\
        & $Pore-HyNet-V_4$(BEST)   & 16 & \textbf{1,692} & \textbf{3.67} & 0.886 & \textbf{93,010} & \textbf{58,162} \\
        & $Pore-SDGMNet-V_4$ & 16 & 927 & 3.50 & \textbf{0.843} & 72,618 & 43,216 \\
        \midrule
        \multirow{3}{*}{\textbf{196}} 
        & $Pore-AFSRNet-V_4$(BEST) & 16 & 4,305 & 4.41 & 0.739 & \textbf{120,777} & 53,960 \\
        & $Pore-HyNet-V_4$   & 16 & \textbf{4,329} & \textbf{4.45} & 0.716 & 112,767 & \textbf{58,907} \\
        & $Pore-SDGMNet-V_4$ & 16 & 3,581 & 4.39 & \textbf{0.694} & 108,981 & 50,861 \\
        \midrule
        \multirow{3}{*}{\textbf{303}} 
        & $Pore-AFSRNet-V_4$ & 16 & 3,647 & 4.10 & 0.830 & \textbf{98,216} & 46,246 \\
        & $Pore-HyNet-V_4$(BEST)   & 16 & \textbf{3,861} & \textbf{4.22} & \textbf{0.826} & 96,868 & \textbf{55,331} \\
        & $Pore-SDGMNet-V_4$ & 16 & 3,574 & 4.12 & 0.827 & 96,173 & 52,970 \\
        \midrule
        \multirow{3}{*}{\textbf{031}} 
        & $Pore-AFSRNet-V_4$(BEST) & 16 & \textbf{4,624} & \textbf{4.83} & 0.718 & \textbf{74,294} & \textbf{90,959} \\
        & $Pore-HyNet-V_4$   & 16 & 4,372 & 4.75 & 0.683 & 71,240 & 78,605 \\
        & $Pore-SDGMNet-V_4$ & 16 & 3,915 & 4.76 & \textbf{0.656} & 74,071 & 65,152 \\
        \midrule
        \multirow{3}{*}{\textbf{038}} 
        & $Pore-AFSRNet-V_4$(BEST) & 16 & \textbf{679} & \textbf{2.95} & 1.204 & \textbf{55,950} & \textbf{28,715} \\
        & $Pore-HyNet-V_4$   & 16 & 509 & 2.90 & 1.234 & 40,018 & 27,585 \\
        & $Pore-SDGMNet-V_4$ & 16 & 541 & 2.83 & \textbf{1.185} & 37,353 & 27,514 \\
        \midrule
        \multirow{3}{*}{\textbf{056}} 
        & $Pore-AFSRNet-V_4$(BEST) & 16 & \textbf{3,625} & 4.14 & 0.717 & \textbf{111,803} & 45,672 \\
        & $Pore-HyNet-V_4$   & 16 & 3,352 & \textbf{4.19} & 0.705 & 96,895 & \textbf{54,453} \\
        & $Pore-SDGMNet-V_4$ & 16 & 2,755 & 4.18 & \textbf{0.644} & 108,077 & 43,285 \\
        \midrule
        \multirow{3}{*}{\textbf{097}} 
        & $Pore-AFSRNet-V_4$(BEST) & 16 & \textbf{820} & 2.99 & 1.151 & \textbf{81,432} & \textbf{24,404} \\
        & $Pore-HyNet-V_4$   & 16 & 644 & \textbf{3.04} & \textbf{1.061} & 61,110 & 22,538 \\
        & $Pore-SDGMNet-V_4$ & 16 & 612 & 2.88 & 1.081 & 58,080 & 20,383 \\
        \midrule
        \multirow{3}{*}{\textbf{All Subjects}} 
        & $Pore-AFSRNet-V_4$(BEST*6) & 16 & \textbf{3,009.9} & 3.799 & 0.904 & \textbf{88,438.0} & 51,593.4 \\
        & $Pore-HyNet-V_4$(BEST*3)   & 16 & 2,912.3 & \textbf{3.866} & 0.877 & 81,943.3 & \textbf{55,074.1} \\
        & $Pore-SDGMNet-V_4$(BEST*0) & 16 & 2,463.4 & 3.788 & \textbf{0.858} & 77,766.1 & 44,911.2 \\
        \bottomrule
    \end{tabular}%
    }
\end{table*}

\subsection{3D Reconstruction on the Final PorePatch Dataset V4}
The final iteration from $PorePatchV_3$ to $PorePatchV_4$ (see Table \ref{tab:recon_v4}) revealed a key phenomenon: performance improvement had reached a saturation point. The substantial performance growth observed in previous iterations did not continue into this stage; for instance, the number of Sparse Points for AFSRNet even slightly decreased in some subject scenes. This trend of diminishing returns was expected and indicates that our iterative process had pushed the dataset's quality and complexity to a high level, where further performance gains for the existing descriptor architectures were marginal. Therefore, the $PorePatchV_4$ dataset represents a well-optimized, final version in this study, at which point we concluded the iterative process. This complete trajectory, from significant improvement to performance saturation, strongly demonstrates the success of our DMCE: it not only effectively boosts model performance but also helps identify a suitable stopping point for the iteration.

On the final $PorePatchV_4$ dataset, we conducted a comprehensive horizontal evaluation of the three leading candidate descriptors: AFSRNet, HyNet, and SDGMNet. The results indicate that AFSRNet delivers the best overall performance. Its advantage lies in generating the highest number of Sparse Points and Dense Points across the majority of subject scenes, while simultaneously maintaining a competitive Mean Track Length and a low Reprojection Error. This ability to maximize model completeness while ensuring high accuracy gives it a significant advantage for practical applications. Meanwhile, HyNet and SDGMNet also demonstrated strong competitiveness, each with distinct strengths: HyNet excelled in reconstruction completeness on specific scenes (e.g., '190', '303'), whereas SDGMNet showcased its exceptional geometric accuracy by frequently achieving the lowest Reprojection Error.

\subsection{Discussion}\label{sec:discussion}
Our extensive experiments provide several key insights. Firstly, the results unequivocally demonstrate the effectiveness of our proposed Data-Model Co-Evolution (DMCE) framework. The iterative process, guided by 3D reconstruction quality, successfully enhanced both the scale and the geometric precision of the dataset, leading to progressively more powerful descriptors. This is evidenced by the steady improvement in reconstruction metrics from $V_1$ to $V_3$.

Secondly, we observed two important phenomena: the ``dip-then-rise'' trend in patch verification and performance saturation in 3D reconstruction. The initial performance dip suggests that the early iterations prioritized dataset expansion and complexity, introducing more challenging samples. The subsequent rise and final saturation indicate that later iterations focused on improving data quality, ultimately reaching a point where the dataset was sufficiently rich and clean for the evaluated models.

Furthermore, we observe that a descriptor's performance in patch verification (e.g., FPR95) is not always strongly correlated with its final performance in the 3D reconstruction task, an observation consistent with findings in related literature \cite{schonberger2017comparative}. This discrepancy arises primarily because patch verification evaluates a descriptor's discriminative power in isolated, pairwise comparisons, whereas 3D reconstruction is a multi-stage, systemic task. 

Finally, the comprehensive benchmark on $PorePatchV_4$ confirms that deep learning descriptors, especially AFSRNet, significantly outperform traditional methods for pore-scale facial feature matching. AFSRNet strikes an optimal balance between the completeness and accuracy of the 3D reconstruction, establishing it as the state-of-the-art for this task. These findings not only validate our dataset and framework but also highlight the remaining challenges and future research directions in handling weak-textured facial regions.

\section{Conclusion}\label{sec:conclusion}

This paper presents a complete solution to address the challenges of local feature matching in weak-textured facial regions and the lack of a dedicated dataset in this domain. We construct a large-scale, pore-level face patch dataset (PorePatch dataset) and pioneer a Data-Model Co-Evolution (DMCE) framework. This framework leverages the best-performing descriptor from each iteration to enhance 3D reconstruction quality, establishing a performance-driven iterative loop  that systematically improves the dataset's quality and challenge.

Within our evaluation benchmark, comprising patch verification and 3D reconstruction, the experimental results clearly demonstrate that: 1) Deep learning-based descriptors far surpass traditional handcrafted methods in handling pore-scale facial features. 2) Our DMCE framework proved highly effective; it not only significantly increases the scale and difficulty of the available data but also enables continuous performance improvement for various descriptors on higher-quality datasets, ultimately reaching saturation with the $PorePatchV_4$. In a comprehensive evaluation, AFSRNet achieves the best balance between the completeness and accuracy of the reconstructed models, exhibiting optimal overall performance. In conclusion, this research not only provides the community with a high-quality, specialized dataset and a comprehensive performance benchmark but, more importantly, our proposed DMCE offers a new paradigm and fresh insights for future research in pore-scale feature learning for specific domains.

\bibliographystyle{bst/sn-vancouver}
\bibliography{mir-reference}

\end{document}